\documentclass[a4, 10 pt, conference]{ieeeconf} % Comment this line out if you need a4paper

\IEEEoverridecommandlockouts % This command is only needed if 
\usepackage[pdftex]{graphicx}
\usepackage{amssymb}
\usepackage{multirow} 

\title{\LARGE \bf
Improving User's Sense of Participation in Robot-Driven Dialogue
}

\author{Makoto Kawamoto$^{1}$ Masaki Shuzo$^{1}$ and Eisaku Maeda$^{1}$% <-this % stops a space
\thanks{*This work was supported by Grant-in-Aid for Scientific Research on Innovative Areas, Grant Numbers JP19H05693.}
\thanks{$^{1}$All authors are with Tokyo Denki University, 5 Senju Asahi-cho, Adachi-ku, Tokyo 120-8551, Japan 
        {\tt\small \{21amj07@ms, shuzo@mail, maeda.e@mail\}.dendai.ac.jp}}
}

\begin{document}

\maketitle
\thispagestyle{empty}
\pagestyle{empty}

\begin{abstract}
In task-oriented dialogues with symbiotic robots, the robot usually takes the initiative in dialogue progression and topic selection. In such robot-driven dialogue, the user's sense of participation in the dialogue is reduced because the degree of freedom in timing and content of speech is limited, and as a result, the user's familiarity with and trust in the robot as a dialogue partner and the level of dialogue satisfaction decrease. In this study, we constructed a travel agent dialogue system focusing on improving the sense of dialogue participation. At the beginning of the dialogue, the robot tells the user the purpose of the upcoming dialogue and indicates that it is responsible for assisting the user in making decisions. In addition, in situations where users were asked to state their preferences, the robot encourages them to express their intentions with actions, as well as spoken language responses. In addition, we attempted to reduce the sense of discomfort felt toward the android robot by devising a timing control for the robot's detailed movements and facial expressions.
\end{abstract}

\section{INTRODUCTION}
In traditional robot dialogue systems, it is sometimes common for the robot to take the initiative in progression and topic selection. We call this as a robot-driven dialogue in this paper, different from human-to-human dialogues in which there is equal initiative each other. In the robot-driven dialogue,
the robot system gives a turn to the user with a question which has some decision 
branches, and wait his/her response. Ordinally, the system cannot accept too long silence and the response while the robot is talking. The system also cannot accept an irregular response that deviates from the current context. While these user's unanticipated responses for the system will be ignored,
next robot's utterance is said to be pre-registered from the previous robot's utterance.

When we have several turns of the robot-driven dialogues, the user tends to have a bad feeling such as tired or disgusting. So, we should maintain user's feeling that he/she wants to keep talking with the robot. We assume that this willingness to dialogue is related to how much he/she join the dialogue during the previous several turns. Although it is difficult to measure this user's sense of participation in the dialogue quantitatively, some information of user's behaviors such as eye gaze, nodding and body orientation will be useful for evaluation.

In the situation where an android robot servers as a sales person at the mock travel agency,
we developed the dialogue system with an experimental participant. Our system based on almost robot-driven dialogues needed some ideas to improve user's sense of participation. In order to increase it, for example, user's hand actions such as pointing in addition to his/her vocal response will be effective. Detail robot system and experimental results are described in this paper.

\begin{figure}[t]
 \centering
 \includegraphics[width=80mm]{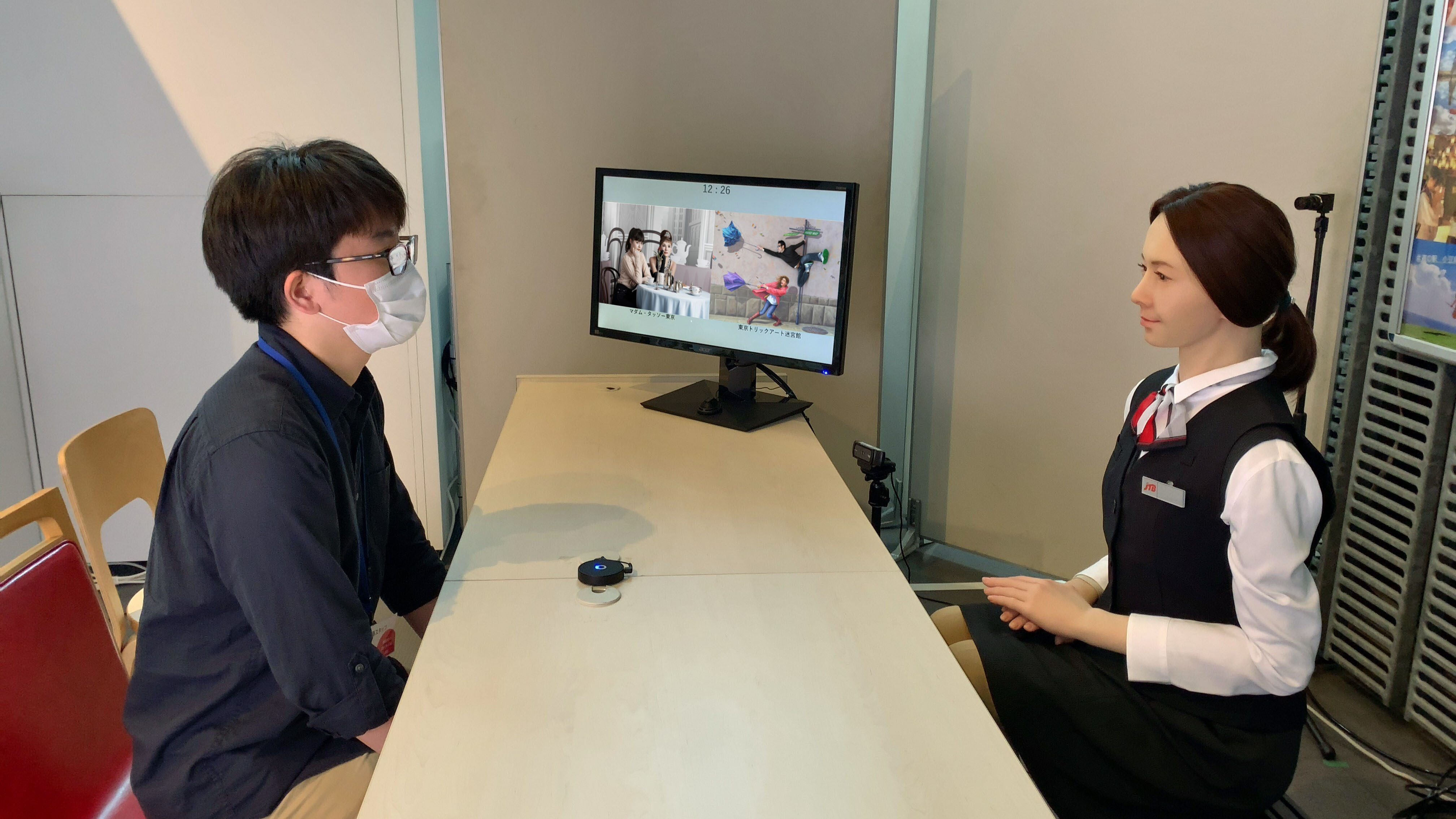}
 \caption{Android I (right) performs dialogue with customer (left). Pictures of two tourist spots chosen by customer are shown on display.}
 \label{figure1}
\end{figure}

\section{DIALOGUE ROBOT COMPETITION}
The first Dialogue Robot Competition (DRC) in 2020 \cite{Higashinaka2022} and the second in 2022 \cite{Minato2022} were held as part of the project of Dialogue Intelligence in the New Academic Research Area. 
% （コンペの体験者は事前に選んだ２つの観光地の内どちらに行くかを決めるためにアンドロイドと対話を行う）アンドロイドI参考文献挿入
In these competitions, a robot (Android I \cite{Nishio 2007}\cite{Glas 2016}) plays the role of a counter salesperson at a travel agency and dialogues with a customer who wants to decide on one travel spot from his/her two pre-selected spots.
DRC's task is to encourage the customer's interest toward the designated spot by the organizers through a 5-minute talk with Android I. 
% The task of Android I is to encourage the customer's interest toward a travel spot designated by the organizers. 
% Hereinafter, this travel spot is referred to as a ``recommended spot.'' 
The dialogue conditions are a one-on-one, face-to-face dialogue between Android I and the customer (Fig. \ref{figure1}). 
The evaluation items in the DRC are based on the customer's satisfaction with the dialogue and whether or not the customer's interest is inclined toward the designated spot. 

\begin{figure}[t]
 \centering
 \includegraphics[scale = 0.8]{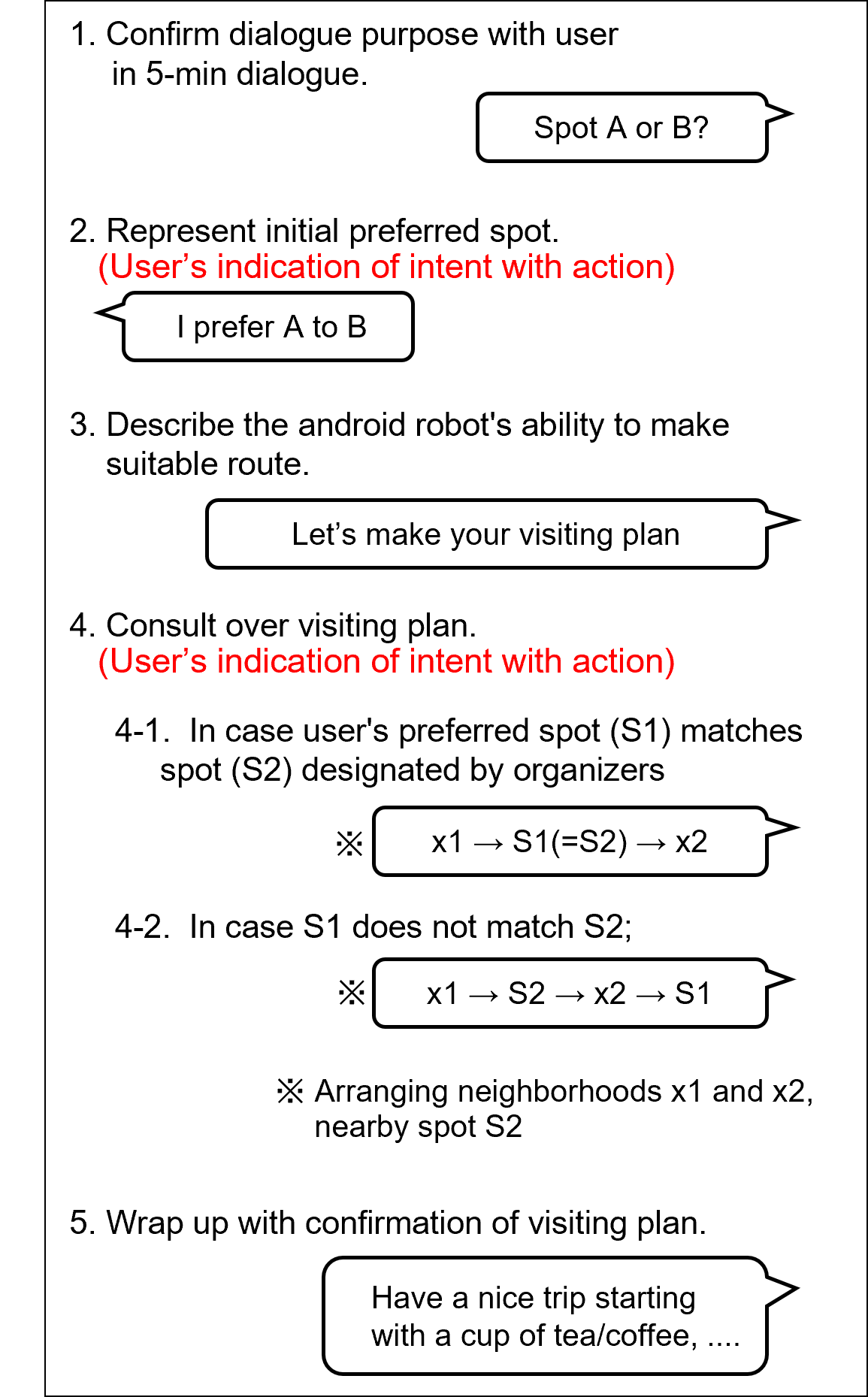}
 \caption{Proposed dialogue flow for making visiting route on basis of user's multimodal input. S1: User's preferred spot, S2: Spot designated by organizers, x1: Neighborhood such as nearby cafe spot S1, and x2: Neighborhood such as nearby restaurant spot S1. }
 \label{figure2}
\end{figure}

\section{DEVELOPMENT OF DIALOGUE SYSTEM TO INCREASE USER'S SENSE OF PARTICIPATION}
Our team, MIYABI, joined the preliminary round of DRC in August 2022. 
The flow of dialogue in our system is shown in Fig. \ref{figure2}. 
The effectiveness of such a sales promotion task using a dialogue agent has been shown in existing studies \cite{Matsumura2017}\cite{Iwamoto2022}. 
The task at DRC could be classified a kind of sales promotion task. Android robot is expected to ensure a certain level of interest from a user (customer). 
Further work is required to improve the user's sense of participation in the dialogue. 
In this section, we describe three methods that we incorporated into this scenario to improve the user's sense of dialogue participation.

\subsection{Clarification of Dialogue Objectives}
The dialogue objective of the user as a visitor to a travel agency is defined as to decide one spot from his/her selected two spots to visit while conversing with the android. The experimental participants in preliminary round of DRC 2022 were not customers in real travel agency but visitors to the mock travel booth set at the National Museum of Emerging Science and Innovation (Miraikan). Before the dialogue session, the participants had been only requested to select two spots from six alternatives where they want to go. However, we could not confirm if the participant really understand the dialogue objective by  the short experimental introduction. If the objective remains some ambiguity, confusing user may lose his/her way to talk with the android.

\begin{figure}[t]
 \centering
 \includegraphics[width=75mm]{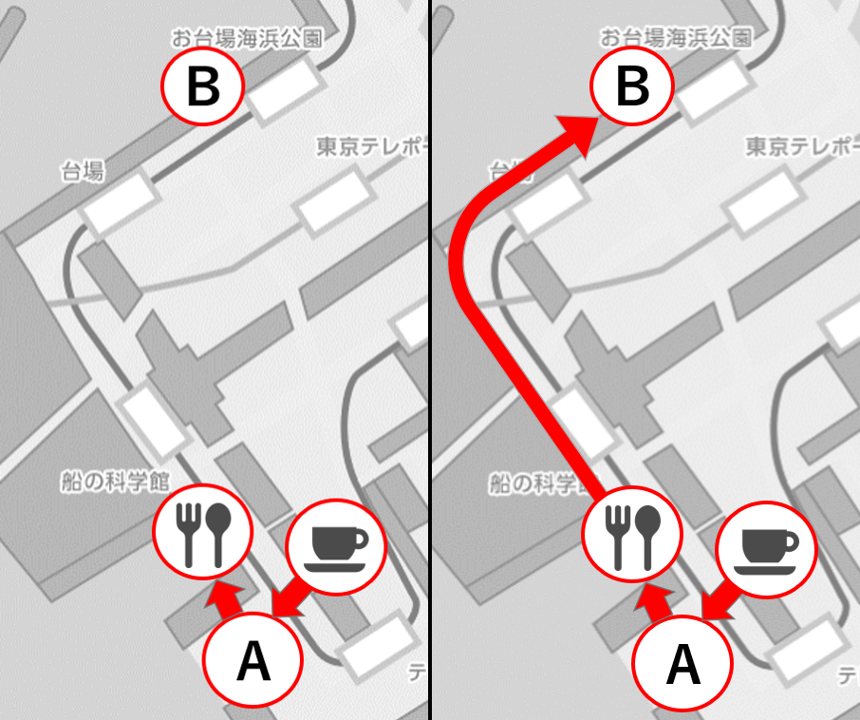}
 \caption{Strategies for route presentation. Left in case where user's preferred spot matches spot designated by organizers; Right in opposite case. A: Tokyo Customs Information Hiroba (designated spot), and B: Tokyo Trick Art Museum as alternative spot.}
 \label{figure3}
\end{figure}

Therefore, at the beginning of the dialogue, the android (not a staff) confirms the objective of the dialogue as followed (Fig. 2). First, the android asks the user, ``Are you sure of which of the two sights you want to visit?'' and he/she replies, ``Yes.'' Then, the android asks, ``Which is your preferred spot?'' and he/she replies ``This spot'' while pointing to the spot's image on the display. Subsequently, the android suggests, ``I like to make a travel plan. 
Let's make your visiting plan together.'' (Strategies for route presentation in Fig. \ref{figure3}) Here, our objective is changed to make the route including the user's preferred spot but is in the category of competition objective.

\subsection{Request to Show Intent Through Actions}
\begin{figure}[t]
 \centering
 \includegraphics[width=\linewidth]{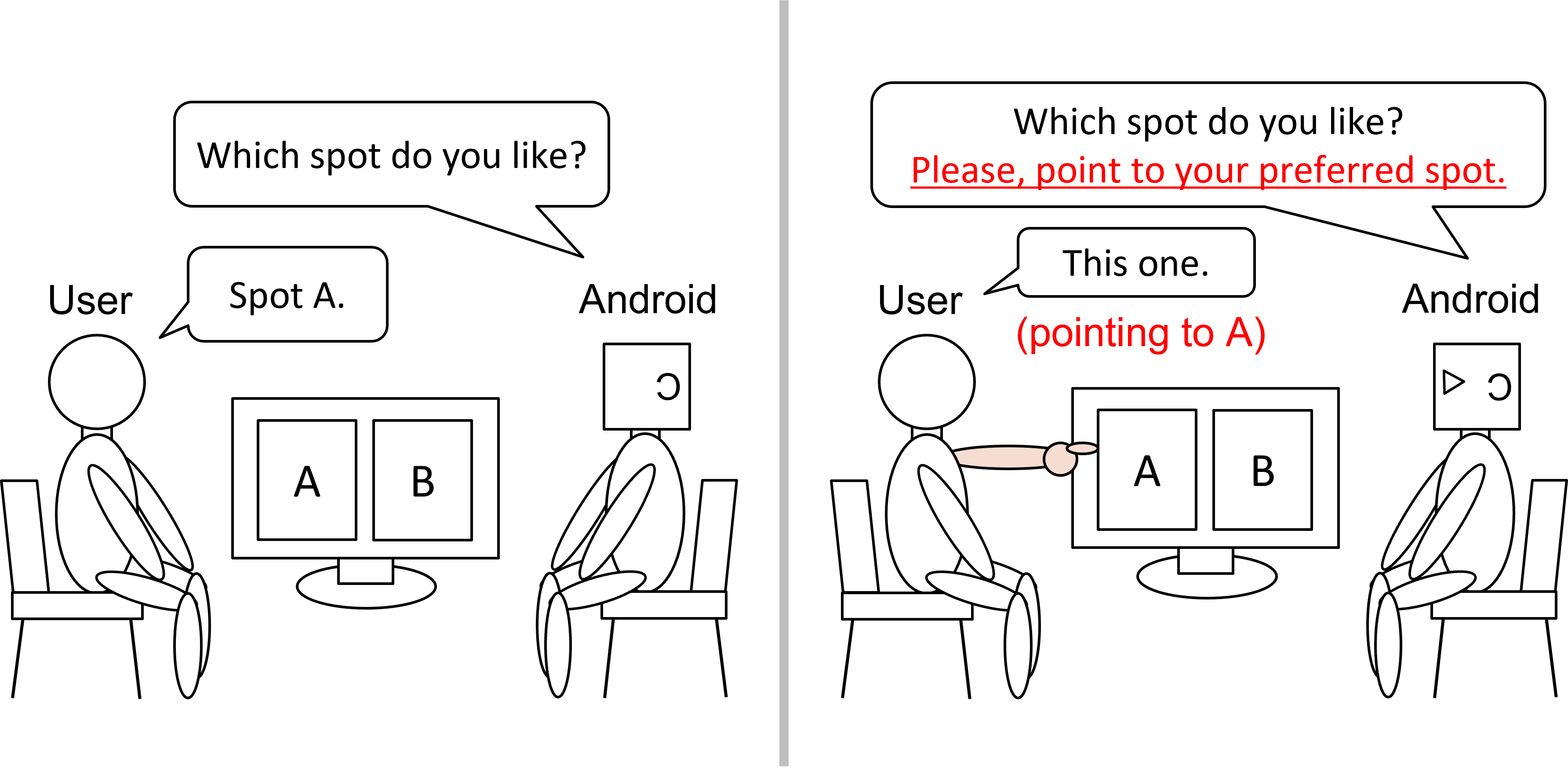}
 \caption{Requesting multimodal behavior for answering robot's question. Left: voice only, Right: voice with gesture.}
 \label{figure4}
\end{figure}

This system incorporates the android's emotional and subjective opinions into the recommendation dialogue. This is not simply a matter of selecting information from a database and providing it to the user but also a matter of creating a friendly relationship between the user and the android by giving the user a sense of the android's character \cite{Watanabe2015}. An android engaging in ``self-talk'' may diminish the user's sense of dialogue participation. Therefore, we asked users interacting with the android to engage in a multimodal dialogue that combines voice input with actions such as pointing and raising hands (Fig. \ref{figure4}). In this way, the user is given the strong impression that the dialogue is shifting in accordance with the user's own choice. In addition, by having the user perform actions toward the android, the user's body is turned toward the android and the display. In this way, we encouraged users whose attention was focused in areas unrelated to the dialogue to concentrate on the dialogue. 
% we infer that the likelihood that the user will reject a request is sufficiently low when the android requests an action from a user.
% When the android requests the user to point or raise his or her hand, it is unlikely that the user will ignore such \cite{Nakagawa2012}\cite{Song2022}.

\subsection{Behavior That Reduces Discomfort}
\begin{figure}[t]
 \centering
 \includegraphics[width=\linewidth]{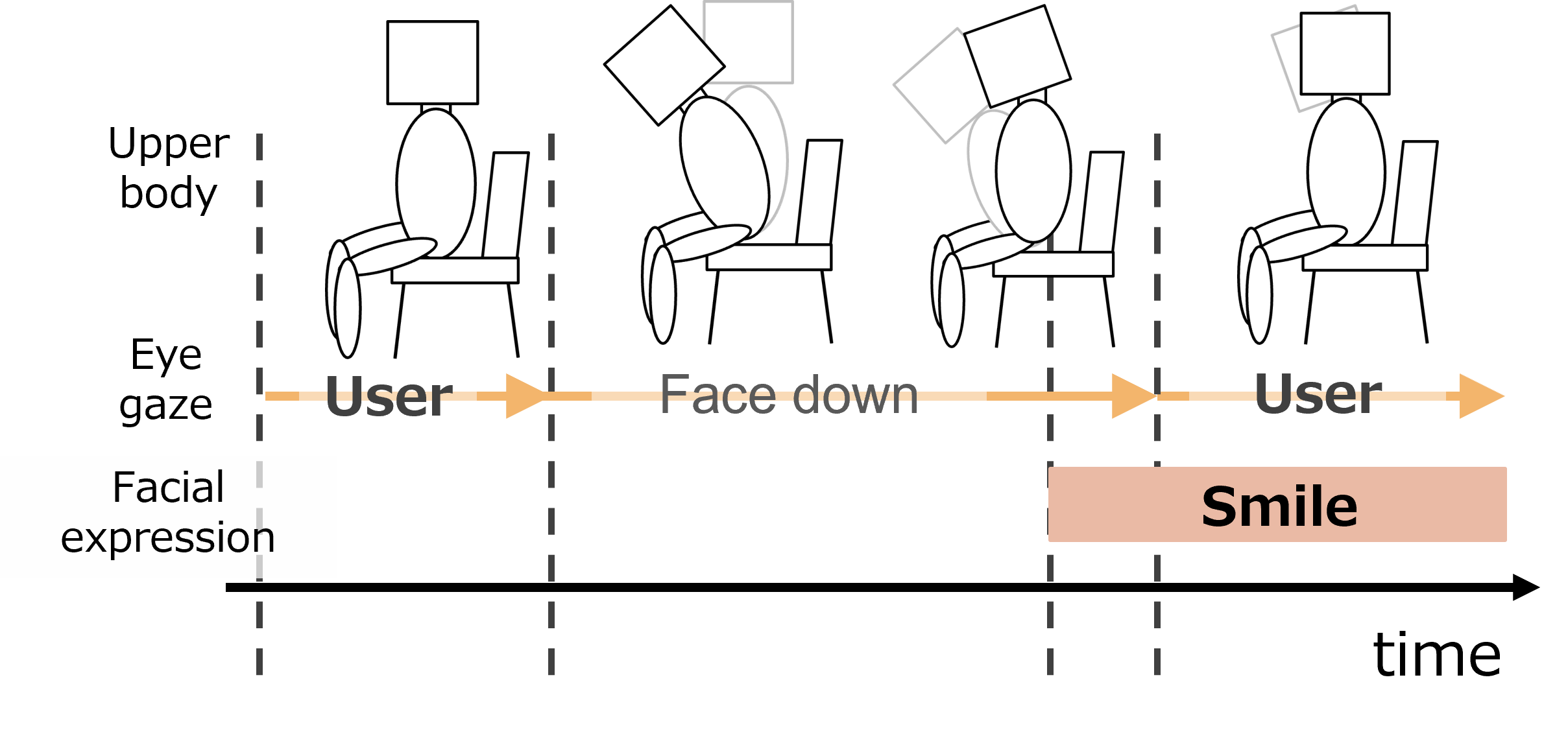}
 \caption{Precise timing control of eye gaze and facial expression for representing politeness in bowing.}
 \label{figure5}
\end{figure}

Because androids closely resemble humans, users demand human-like behavior from them \cite{Chaminade2006}. Therefore, we incorporated ideas to eliminate the android's robotic nature. First, ``bowing,'' an important behavior in a travel agency in countries such as Japan, was represented by the movements of the android's upper body, gaze, and facial expressions. Proper bowing increases the performer's attractiveness and gives the impression of politeness and obedience to others. In this system, the android was made to initiate bowing after the greeting utterance was completed. To further enhance the politeness of the gesture, the android robot took a short pause after raising its upper body and then slowly turned its gaze toward the user. In this case, by giving a gap between the android's movements and eye movements, a more polite and human-like impression was given to the user. In addition, to make the transitions in facial expression natural, the android was made to smile even before the bowing motion was completed. The transitions of the upper body, gaze, and facial expression during bowing are shown in Fig. \ref{figure5}.

\begin{figure}[t]
 \centering
 \includegraphics[width=\linewidth]{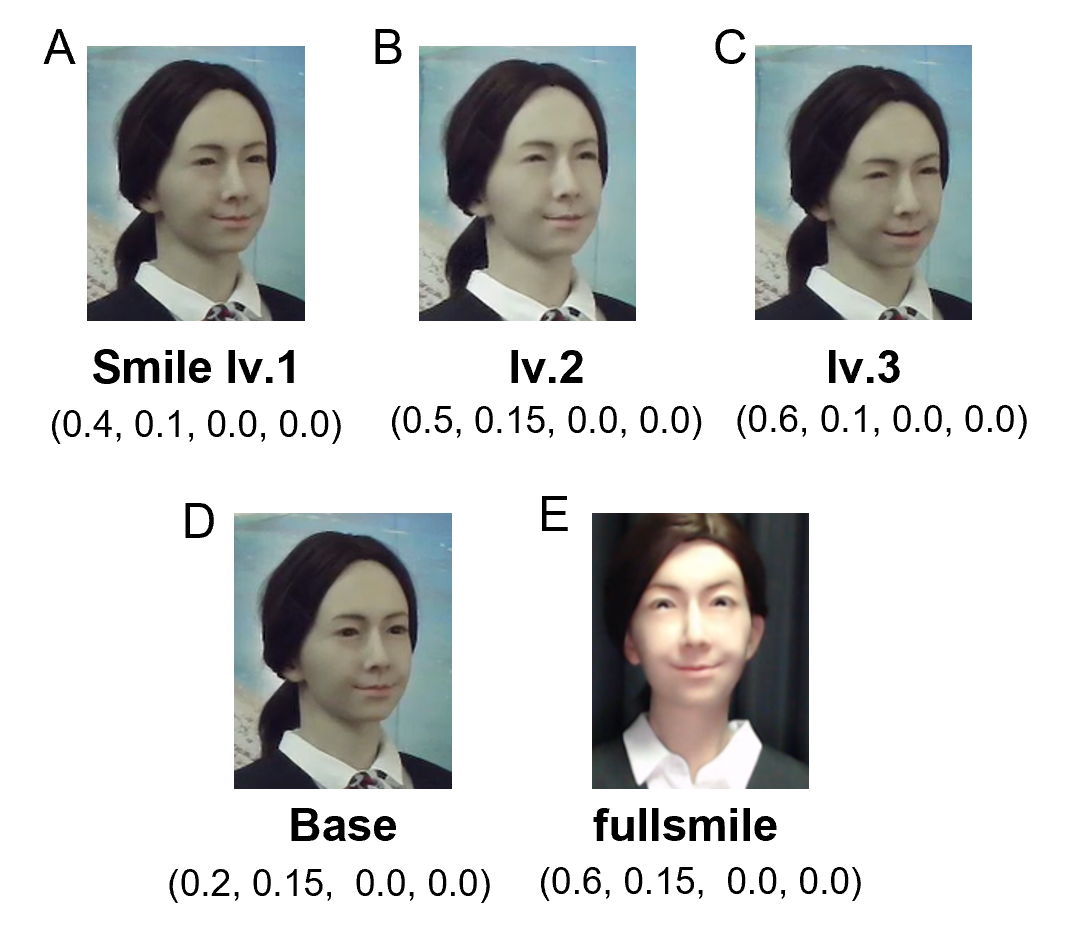}
 \caption{Three types of natural smiles (A--C) that we made using basic facial expression (D) of Android I. Fullsmile (E) provided by organizers was not used. Numbers in parentheses are expression parameters in application JointMapperPlusUltraSuperFace used to create facial expressions (valence, arousal, dominance, real intention).}
 \label{figure6}
\end{figure}

Second, smiling is an effective way to make androids more human \cite{Chikaraishi2008}. We therefore designed a natural smile for the android. The designed facial expressions are shown in Fig. \ref{figure6}. Users may not be able to recognize slight changes in an android's facial expression, so in addition to the naturalness and impression of the android's smile itself, this system was designed to make the android's subsequent smile stand out in contrast by moderating the degree to which the corners of its mouth are lifted and the degree to which its eyes are narrowed immediately before its smile appears.

\begin{figure}[t]
 \centering
 \includegraphics[width=60mm]{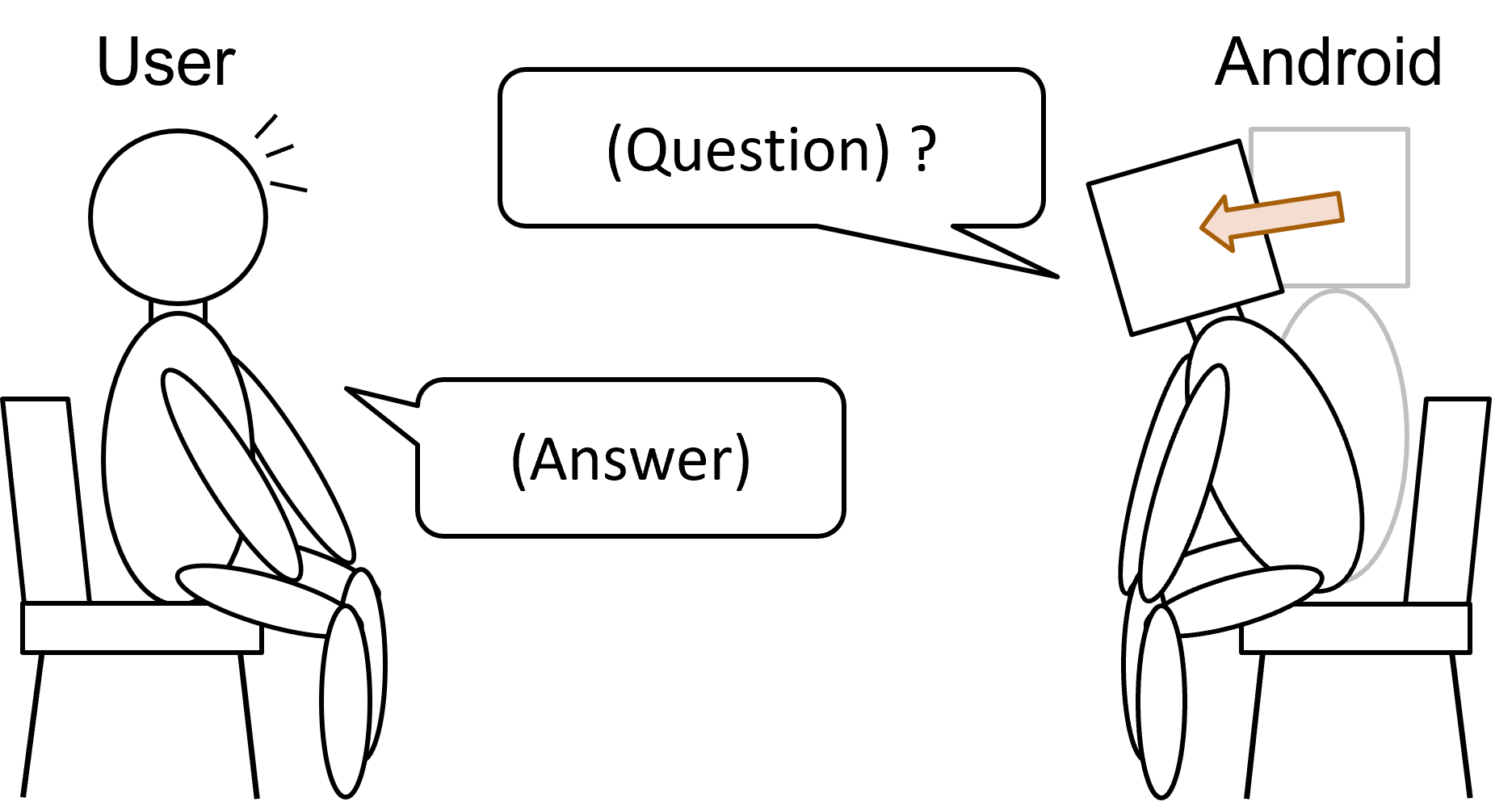}
 \caption{Indication of android intent through subtle behavior. Forward leaning posture shows intention to give utterance turn.}
 \label{figure7}
\end{figure}

Third, to facilitate turn-taking from the android to the user, we implemented a behavior in the android that makes it always lean forward when asking a question to the user (Fig. \ref{figure7}). In this way, it becomes easier for the user to grasp the timing of his or her own speech. Even if the user misses the utterance of the android, the fact that the current turn of utterance is the android's is conveyed to the user, allowing the user to take the action of listening back.

\section{RESULT AND DISCUSSION}
The results of a questionnaire evaluation done in the preliminary round of DRC 2022 were used to confirm the effectiveness of the three methods. In addition, to confirm whether a participant was immersed in the dialogue, we analyzed his/her gaze and behavior while the android spoke.

The results of the evaluation are shown in Table 1. The proposed system outperformed the baseline system in terms of satisfaction with choice (Sat/c), $4.72 \pm 1.93 > 4.19$. It also obtained a higher rating than the baseline system for dialogue favorability (Lik), $5.07 \pm 1.78 > 4.59$. Dialogue satisfaction (Sat/d) was also higher than the baseline at $4.72 \pm 1.96 > 4.15$. Thus, the results suggest that the proposed system was favorably accepted by the participants. Statistical tests is not shown because individual results of other teams are not opened by the organizers.

Regarding the fact that participants were asked to express their intentions through actions, 27 out of 29 participants responded when the android asked for input through actions. At that time, the participants turned their bodies toward the android and the display. They tended to continue looking at the android even after the input was completed. 

\addtolength{\textheight}{-8.82cm}

% companion
However, during the android's speech, the participant's gaze frequently shifted toward his/her companions sitting next to the participant (his/her family, friends, etc.) as well as the android and the display. Also, 
before his/her speech responding to robot's question,
% during voice input,
participant consulted with his/her companions 
how to reply.
% about the content of the response before speaking. 
This meant that the expected one-on-one dialogue did not take place, but rather, participant-to-android and participant-to-companions were engaged in dialogue at the same time. 

\begin{table}[ht]
\centering
\caption{Results of participant's impression rating of android in preliminary round of Dialogue Robot Competition. 
``Baseline system'' is general recommendation dialogue system created by organizers.
``All entry system'' is scored on average survey results of participating teams (12 teams in all). 
sat/c, inf, nat, app, lik, sat/d, tru, use, and reu denote satisfaction with choice, informativeness, naturalness, appropriateness, likeability, satisfaction with dialogue, trustworthiness, usefulness, and intention to reuse, respectively.}
\begin{tabular}{ccccl}
\cline{1-4}
\multirow{2}{*}{\begin{tabular}[c]{@{}c@{}}Evaluation\\ item\end{tabular}} & \begin{tabular}[c]{@{}c@{}}Proposal\\ system\end{tabular} & \begin{tabular}[c]{@{}c@{}}Baseline\\ system\end{tabular} & \begin{tabular}[c]{@{}c@{}}All entry \\ system\end{tabular} & \\
 & mean±SD & mean & mean & \\ \cline{1-4}
Sat/c & 4.72±1.93 & 4.19 & 4.48 & \\
Inf & 4.89±1.63 & 3.96 & 4.46 & \\
Nat & 3.76±1.91 & 3.81 & 3.74 & \\
App & 4.24±1.98 & 4.41 & 4.32 & \\
Lik & 5.07±1.78 & 4.59 & 4.60 & \\
Sat/d & 4.72±1.96 & 4.15 & 4.45 & \\
Tru & 4.34±1.92 & 4.30 & 4.35 & \\
Use & 5.17±1.58 & 4.67 & 4.83 & \\
Reu & 4.52±2.14 & 4.07 & 4.23 & \\ \cline{1-4}
\end{tabular}
\end{table}

\section{CONCLUSION}
In this paper, we examined the android's facial expressions, speech, and movements that were incorporated to increase the user's sense of participation in a robot-driven dialogue. 
It was suggested that clarifying the purpose of the dialogue and setting a more specific purpose, in this case, creating a travel route, so that the user can be more immersed in the dialogue will improve the user's comprehension of the information. 
In addition, it is suggested that having the user indicate his/her intent through actions, though reducing the naturalness of the dialogue compared with dialogue done through voice input alone, increases the user's attention to the android and improves the user's concentration on the subsequent dialogue. 
The results also suggest that the android's behavior, designed to reduce the chance of the android making the user feel uncomfortable, was effective in facilitating the smooth progress of the dialogue. 
Future work will examine ways to incorporate these methods while ensuring the naturalness of the dialogue. 
% 本稿では，ロボット主導型対話におけるユーザの参与意識を高めるために取り入れた，アンドロイドの表情や発話，動作等に関する工夫について検討を行った，対話目的の明確化を行うことや，ユーザがより対話に没入できるよう，旅行ルートを作成するというより具体的な対話目的を新たに設定し対話を実施したことは，ユーザの情報理解度を向上させることが示唆された．また，動作による意思表示をユーザに行わせることは，音声入力のみの対話と比較して、対話の自然さは損なうものの，アンドロイドロボットへの注目度を高めその後の対話への集中度を向上させることが示唆された．またアンドロイドの違和感を低減させる振る舞いは対話の円滑な進行に有効であったことが示唆された．今後は対話の自然さを確保しながらこれらの工夫を取り入れる方法について検討を行う．\\

\end{document}